\documentclass{article}

    \usepackage[final,nonatbib]{neurips_2022}

\usepackage[utf8]{inputenc} 
\usepackage[T1]{fontenc}    
\usepackage{hyperref}       
\usepackage{url}            
\usepackage{booktabs}       
\usepackage{amsfonts}       
\usepackage{nicefrac}       
\usepackage{microtype}      
\usepackage{xcolor}         
\usepackage{svg}
\usepackage{amsmath}
\usepackage{caption}
\usepackage{subcaption}

\title{Towards Real-Time Text2Video via CLIP-Guided, Pixel-Level Optimization}

\author{%
  Peter Schaldenbrand\thanks{ The Robotics Institute, Carnegie Mellon University. \{pschalde,zhixuan2,hyaejino\}@andrew.cmu.edu} \\
   \And
   Zhixuan Liu\footnotemark[1] \\
   \And
   Jean Oh\footnotemark[1] \\
}

\begin{document}

\maketitle
\vspace{-10pt}
\begin{abstract}\vspace{-3pt}
  We introduce an approach to generating videos based on a series of given language descriptions. 
  Frames of the video are generated sequentially and optimized by guidance from the CLIP image-text encoder; iterating through language descriptions, weighting the current description higher than others.
  As opposed to optimizing through an image generator model itself, which tends to be computationally heavy, the proposed approach computes the CLIP loss directly at the pixel level, achieving general content at a speed suitable for near real-time systems.
  The approach can generate videos in up to 720p resolution, variable frame-rates, and arbitrary aspect ratios at a rate of 1-2 frames per second.
  Please visit our website to view videos and access our open-source code: \url{https://pschaldenbrand.github.io/text2video/}.
\end{abstract}
\vspace{-7pt}

\begin{figure}[!hb]
    \centering \scriptsize
    \includegraphics[width=\textwidth]{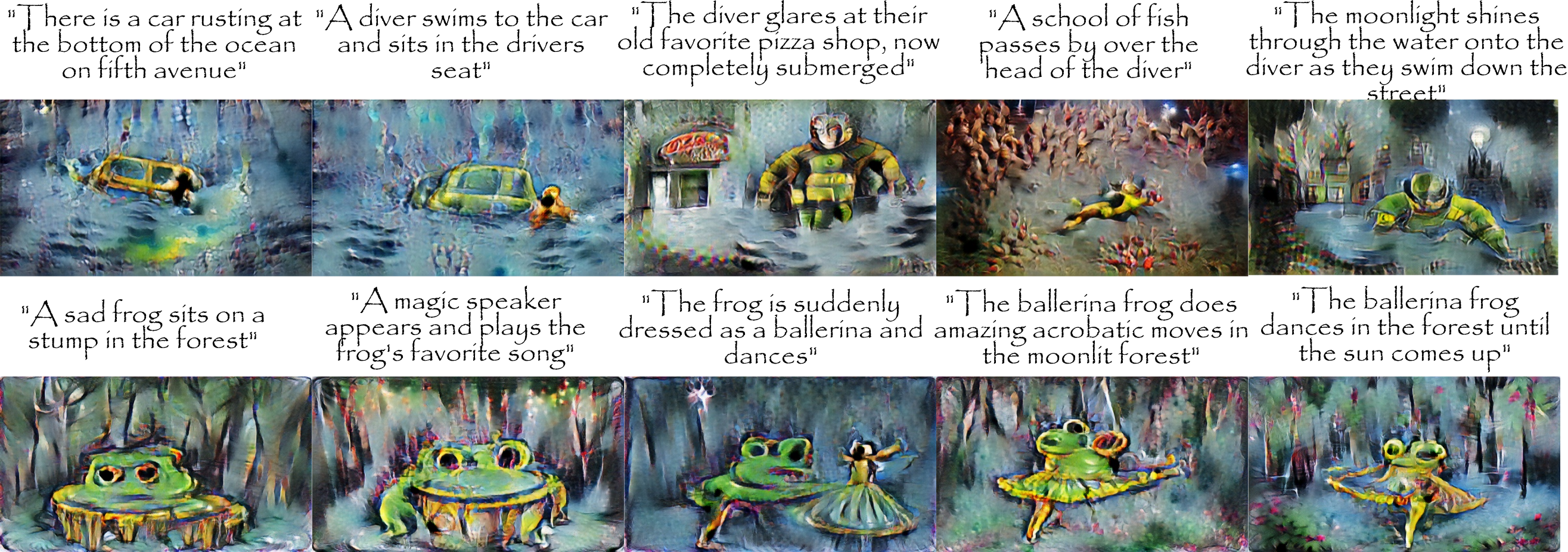}
    \caption{ Frames sampled from videos generated by our approach produced with a series of language descriptions expressing narratives.}
    \label{fig:my_label}
\end{figure}\vspace{-3pt}
\textbf{Introduction}
Animation is a compelling medium allowing unlimited degrees of visual expression while requiring its creator both the artistic skills and a tremendous amount of time to produce an artifact. 
Many people do not have the time and resources to learn animation software or create by hand; however, most can describe the visual elements of a story with words. To bridge this gap, we introduce a method for generating animated videos using natural language input.

Existing CLIP-Guided Text2Video and Text2Image approaches utilize pretrained image generator models, such as Diffusion. Using a pretrained image generator model ensures that the output will appear in distribution to the generator's training data.  While this can produce highly-realistic imagery, it constrains the variety of producible content.  
Optimizing through large generator models is also time consuming. A single frame of video generated with Disco Diffusion \cite{crowson2021disco} takes on the order of 5 minutes, 17 seconds for animation adaptations of Stable Diffusion \cite{rombach2022high}, and 1 minute for CogVideo \cite{hong2022cogvideo}  while our model-free approach generates 1-2 frames per second.



\textbf{Approach}
To achieve real-time text2video generation, we propose a two-step approach; (1) quickly and noisily generating semantic content, then (2) refining image textures in a post-processing step.

\begin{figure}
\centering
\begin{minipage}{.4\textwidth}
  \centering \small
    \includegraphics[width=5.5cm]{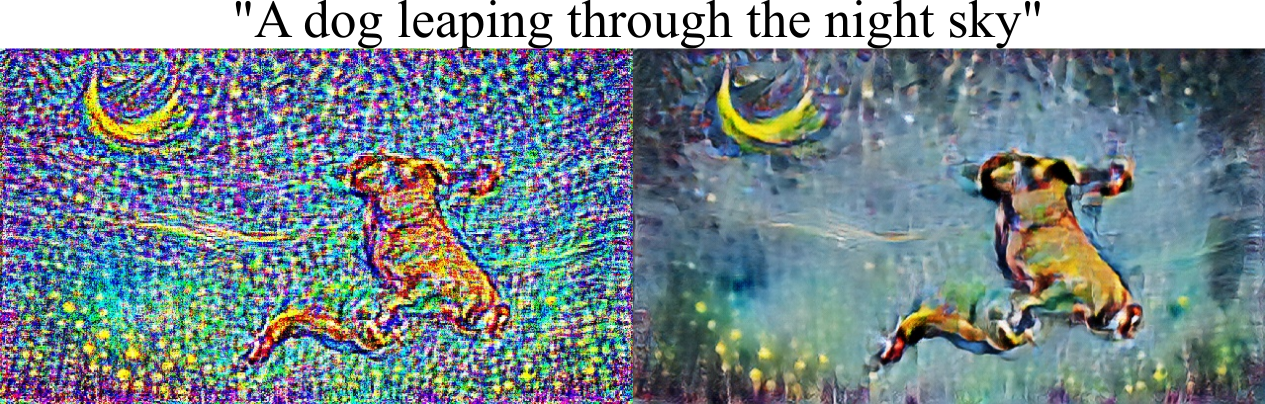}%
  \captionof{figure}{\small Our trained CycleGAN~\cite{zhu2017cyclegan} model denoising the directly optimized  pixels (left).}
  \label{fig:cyclegan}%
\end{minipage}%
\begin{minipage}{.6\textwidth}
  \begin{align}\vspace{-20pt}
    \begin{split}
        \mathcal{L}_{text} &= w_n \cos(E(\texttt{frame}_t), E(\texttt{prompt}_n)) \\ 
        & + w_{n+1} \cos(E(\texttt{frame}_{t}), E(\texttt{prompt}_{n+1})) \label{eq:text_align}
    \end{split}\\
    \mathcal{L}_{stable} &= w_c || \texttt{frame}_t - \texttt{frame}_{t-1} ||_1 \label{eq:img_consistence}
\end{align}
\begin{align}\small
    \min_{\texttt{frame}_t} & \mathcal{L}_{text} + \mathcal{L}_{stable} \label{eq:objective}
\end{align}
\end{minipage}%
\end{figure}

To generate the semantic content, we alter the pixels of the video frame directly to decrease the difference between the frame and the text prompts.
We generate each frame sequentially while iterating through given text prompts to guide the content. 
The first frame is initialized with uniformly distributed noise. We draw on past CLIP-Guided techniques~\cite{frans2021-clipdraw,schaldenbrand2022styleclipdraw,Galatolo2021-ClipStyleGAn2,smith2021-clipGuided} to compare the frame and the language description: the augmented frames and the text prompt are encoded using CLIP~\cite{radford2021-clip} ($E$) and compared using cosine distance (Eq. \ref{eq:text_align}).  For the first frame, only the first prompt is used, and for subsequent frames, a linearly interpolated weight determines how much each prompt influences the generated frame.
The initial state of each frame past the first is the prior frame plus some noise. Using the prior frame encourages frame to frame consistency of the locations of the content, while adding noise ensures that the frames do not get stuck in local minima and some modifications are made between frames.  An extra measure to ensure the frames do not change too drastically is adding a video stability loss (Eq.~\ref{eq:img_consistence}) with a small amount of weight.

Our objective function (Eq.~\ref{eq:objective}) encourages semantic content generation but does not influence the appearance or texture producing images that are very noisy in appearance. We train CycleGAN~\cite{zhu2017cyclegan} to denoise these images, training the model to translate from noisy video frames to photographs~\cite{plummer2015flickr30k}. Fig.~\ref{fig:cyclegan} shows examples of the translation. The frames processed with CycleGAN do not appear photographic, but are smoother and have more realistic colors.

A user can specify the number of frames for each text prompt they give. A temperature parameter controls strength of the video stability loss (Eq.~\ref{eq:img_consistence}) and the amount of noise to add between frames.

\begin{figure}[t]
    \centering
    \includegraphics[width=\textwidth]{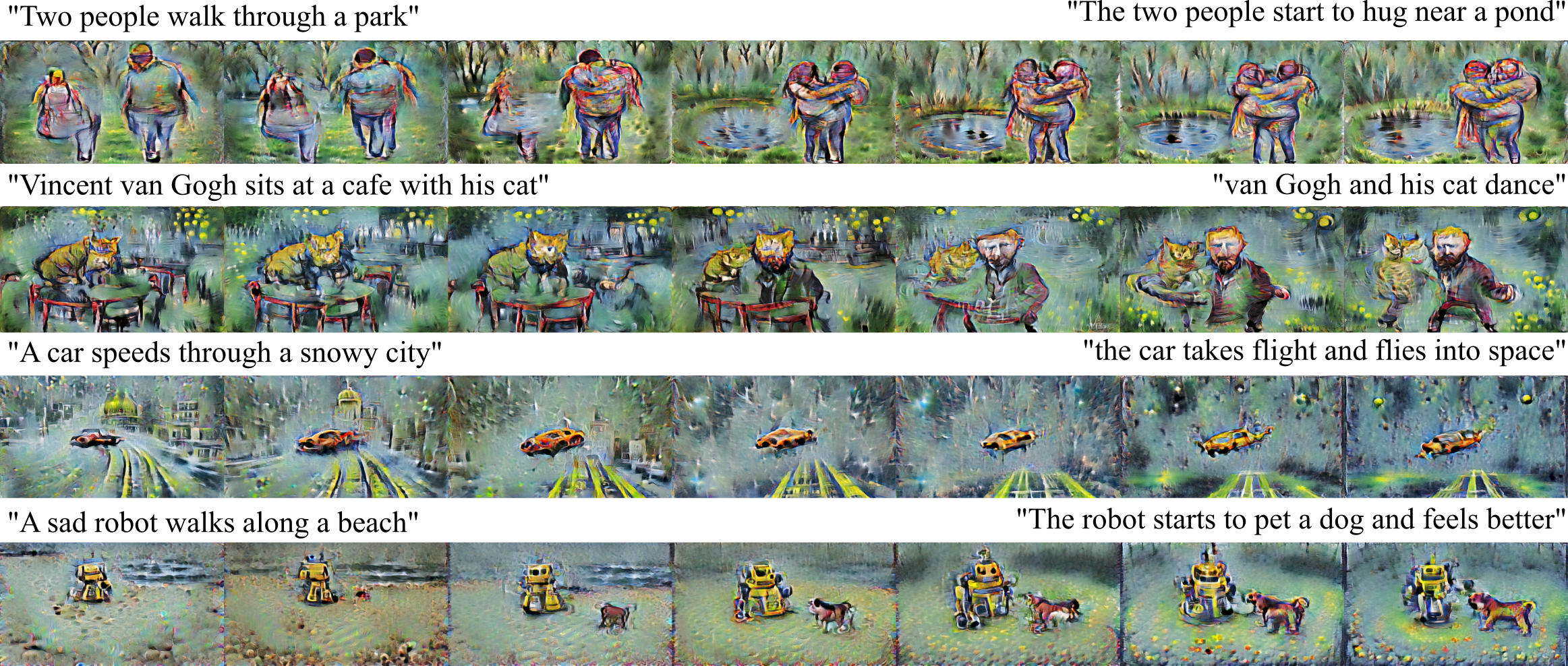}
    \caption{
    Samples from 60 frame videos generated using two text prompts, each in $\sim100$ seconds.}
    \label{fig:two_prompts}
\end{figure}

\textbf{Results}
We demonstrate that 
our approach can generate images of general concepts in different configurations/combinations as shown in Fig.~\ref{fig:two_prompts} 
with more results at: \url{https://pschaldenbrand.github.io/text2video/}.
Empirically, our approach is 20-300 times faster than existing approaches~\cite{crowson2021disco,rombach2022high,hong2022cogvideo}.

\textbf{Discussions}
The appearance of the videos can be noisy and bizarre and is currently beyond user control. 
In future work, we intend to add priors to ensure smoother motion in the videos and more user control over the appearance and style of the videos.

\textbf{Ethical Considerations}
Visual content generation is controlled by CLIP~\cite{radford2021-clip} in this approach. CLIP is known to carry over biases from its training data which reflect many wrong and harmful ideals. It is important that the user of this work understands the biases and tendencies of CLIP. It is also important that users do not use this work to create misinformation or harmful content.

\textbf{Acknowledgement}
This work was supported in part by NSF IIS-2112633 and the Technology Innovation Program (20018295, Meta-human: a virtual cooperation platform for a specialized industrial services) funded By the Ministry of Trade, Industry \& Energy(MOTIE, Korea).
\vspace{-7pt}
\bibliographystyle{plain}
\bibliography{ref.bib}

\begin{thebibliography}{10}

\bibitem{crowson2021disco}
Katherine~Crowson et~al.
\newblock Disco diffusion.
\newblock \url{http://discodiffusion.com/}, 2021.

\bibitem{frans2021-clipdraw}
Kevin Frans, LB~Soros, and Olaf Witkowski.
\newblock Clipdraw: Exploring text-to-drawing synthesis through language-image
  encoders.
\newblock {\em arXiv preprint arXiv:2106.14843}, 2021.

\bibitem{Galatolo2021-ClipStyleGAn2}
Federico Galatolo., Mario Cimino., and Gigliola Vaglini.
\newblock Generating images from caption and vice versa via clip-guided
  generative latent space search.
\newblock {\em Proceedings of the International Conference on Image Processing
  and Vision Engineering}, 2021.

\bibitem{hong2022cogvideo}
Wenyi Hong, Ming Ding, Wendi Zheng, Xinghan Liu, and Jie Tang.
\newblock Cogvideo: Large-scale pretraining for text-to-video generation via
  transformers.
\newblock {\em arXiv preprint arXiv:2205.15868}, 2022.

\bibitem{plummer2015flickr30k}
Bryan~A Plummer, Liwei Wang, Chris~M Cervantes, Juan~C Caicedo, Julia
  Hockenmaier, and Svetlana Lazebnik.
\newblock Flickr30k entities: Collecting region-to-phrase correspondences for
  richer image-to-sentence models.
\newblock In {\em Proceedings of the IEEE international conference on computer
  vision}, pages 2641--2649, 2015.

\bibitem{radford2021-clip}
Alec Radford, Jong~Wook Kim, Chris Hallacy, Aditya Ramesh, Gabriel Goh,
  Sandhini Agarwal, Girish Sastry, Amanda Askell, Pamela Mishkin, Jack Clark,
  Gretchen Krueger, and Ilya Sutskever.
\newblock Learning transferable visual models from natural language
  supervision.
\newblock {\em CoRR}, abs/2103.00020, 2021.

\bibitem{rombach2022high}
Robin Rombach, Andreas Blattmann, Dominik Lorenz, Patrick Esser, and Bj{\"o}rn
  Ommer.
\newblock High-resolution image synthesis with latent diffusion models.
\newblock In {\em Proceedings of the IEEE/CVF Conference on Computer Vision and
  Pattern Recognition}, pages 10684--10695, 2022.

\bibitem{schaldenbrand2022styleclipdraw}
Peter Schaldenbrand, Zhixuan Liu, and Jean Oh.
\newblock Styleclipdraw: Coupling content and style in text-to-drawing
  translation.
\newblock In {\em Proceedings of the International Joint Conference on
  Artificial Intelligence}, 2022.

\bibitem{smith2021-clipGuided}
Amy Smith and Simon Colton.
\newblock Clip-guided gan image generation: An artistic exploration.
\newblock {\em Evo* 2021}, page~17, 2021.

\bibitem{zhu2017cyclegan}
Jun-Yan Zhu, Taesung Park, Phillip Isola, and Alexei~A Efros.
\newblock Unpaired image-to-image translation using cycle-consistent
  adversarial networks.
\newblock In {\em Proceedings of the IEEE international conference on computer
  vision}, pages 2223--2232, 2017.

\end{thebibliography}

\appendix
\section{Temperature Parameter}
\begin{figure}
    \centering
    \includegraphics[width=\textwidth]{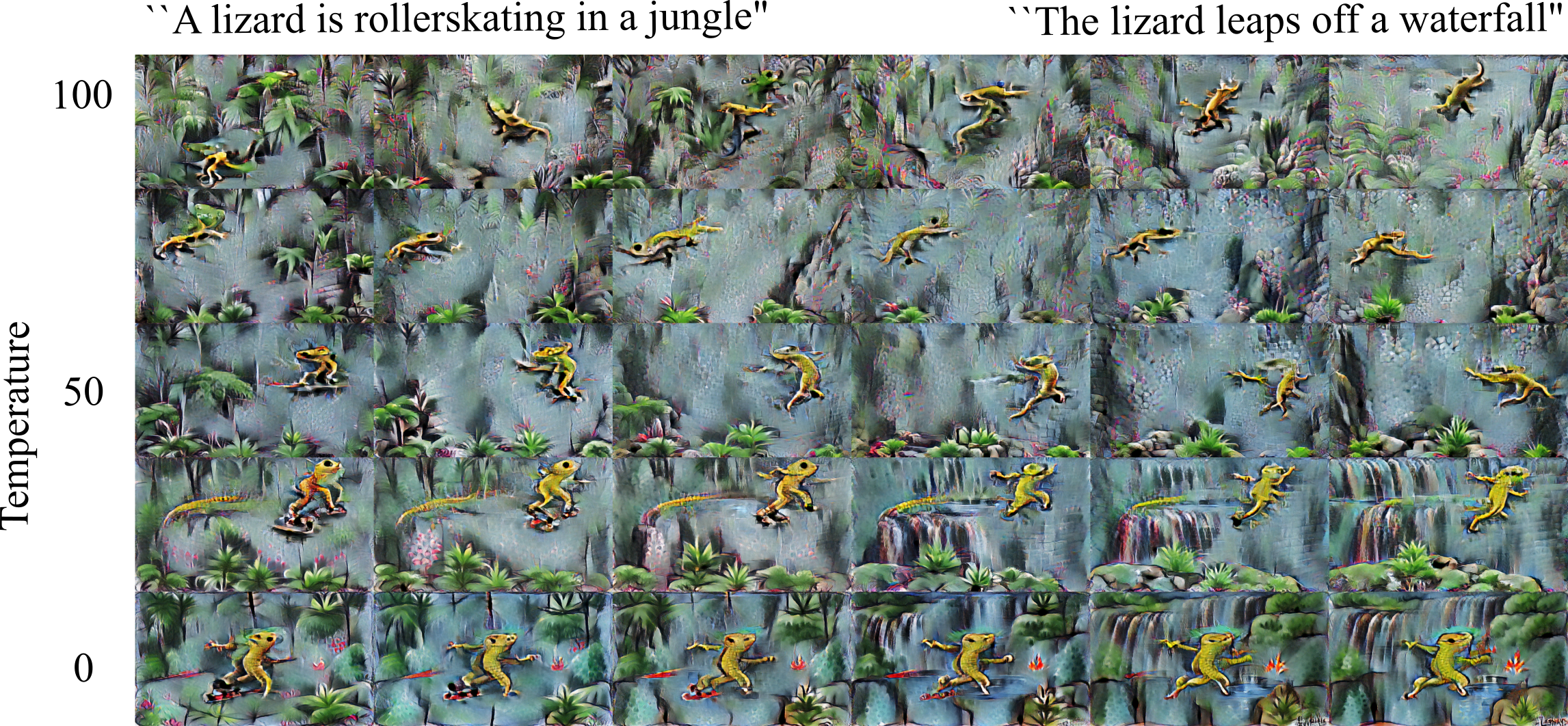}
    \caption{\small Sampling 6 (out of 60) frames from generated videos using two input text prompts and varying the temperature parameter.}
    \label{fig:temp_vs_time}
\end{figure}

\vspace{-12pt}
We explored the effects of the temperature parameter (Fig.~\ref{fig:temp_vs_time}).
Temperature is a value between 0 and 100 that users specify to control how much frame-to-frame differences should be encouraged. Temperature controls the weight of the video stability loss ($w_c$ in Eq.~\ref{eq:img_consistence}) and the standard deviation of the noise added to frames when initializing the subsequent frame with the previous.
The exact relationship between temperature and these parameters was hand written based on observations of the effects of the parameters.
Objects and settings with low temperature barely move frame-to-frame, while with high temperature the scene completely changes (Fig.~\ref{fig:temp_vs_time}).

\section{Resolution and Aspect Ratio}
\begin{figure}
    \centering
    \includegraphics[width=\textwidth]{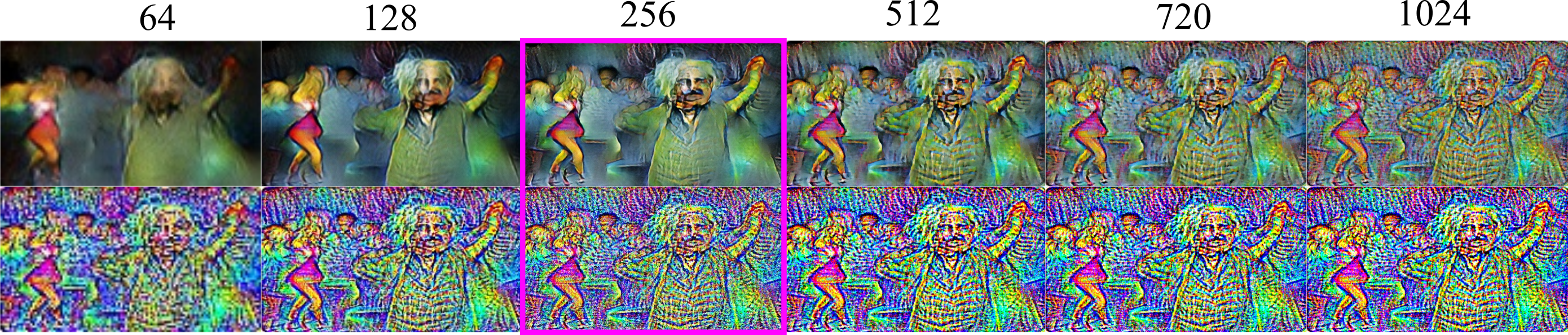}
    \caption{\small A video frame with a vertical height of 256 pixels was generated with the language description ``Albert Einstein dancing in the club". The image, prior to post-processing with CycleGAN, was scaled to different resolutions to investigate CycleGAN's denoising abilities on different sized images.}
    \label{fig:diff_res_cyclegan}
\end{figure}

\vspace{-12pt}
A user specifies the resolution of the generated video a priori.  CLIP requires $224\times224$ resolution input images, however, with the augmentation step, video frames can be cropped and resized to fulfill this requirement.  Our approach operates with arbitrary resolutions, but the resolution does impact both the content and the appearance of the generated video frames. Lower resolutions produce simpler and smoother video frames, while a high resolutions are noisy, sparse, and do not align to the text prompt as well as medium resolutions (Fig.~\ref{fig:temp_vs_time}).  To investigate our post-processing model's involvement in the altering of the appearance of generated frames at different resolutions, we resized the same image to multiple resolutions prior to post-processing (Fig.~\ref{fig:diff_res_cyclegan}). At low resolutions, our post processing model, CycleGAN, smooths the images greatly but has little effect at denoising with high resolutions.

\begin{figure}
    \centering
    \includegraphics[width=\textwidth]{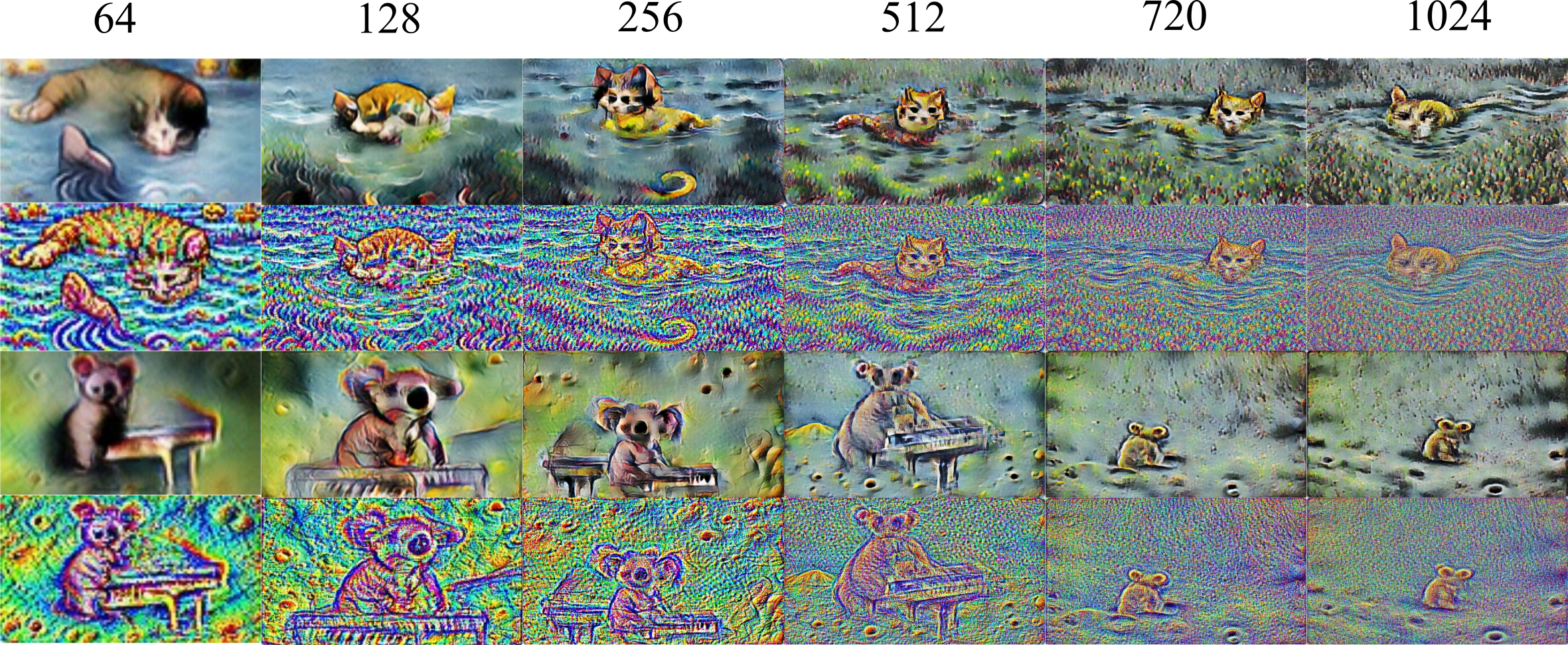}
    \caption{\small Generating video frames with varying resolutions. 
    The vertical pixel resolution is shown above each generated frame with the post-processed frame displayed above the pixels that were directly optimized. The language prompts used: (top) ``A cat swimming in the ocean" and (bottom) ``A koala playing the piano on mars".}
    \label{fig:resolutions}
\end{figure}

\end{document}